\documentclass[10pt,twocolumn,letterpaper]{article}

\usepackage[pagenumbers]{cvpr}
\usepackage{booktabs}
\usepackage{multirow}
\usepackage[table]{xcolor}
\usepackage{graphicx}
\usepackage[most]{tcolorbox}
\usepackage{subcaption}
\usepackage{tabularx}
\usepackage{array}

\newcolumntype{Y}{>{\centering\arraybackslash}X}
\definecolor{cvprblue}{rgb}{0.21,0.49,0.74}
\definecolor{darkgreen}{HTML}{0c8918}
\definecolor{darkgrey}{rgb}{0.53,0.53,0.53}
\definecolor{mygrey}{rgb}{0.9,0.9,0.9}

\usepackage[pagebackref,breaklinks,colorlinks,allcolors=cvprblue]{hyperref}

\def\confName{CVPR}
\def\confYear{2027}

\title{ActField: Learning Action Effects as Continuous-Time Latent Dynamics}

\author{
Yunpeng Qing$^{1,*}$ \quad
Yuheng Huang$^{1,*}$ \quad
Changxu Wei$^{2}$ \quad
Yilun Kong$^{3}$ \\
Yichen Shi$^{4}$ \quad
Shunyu Liu$^{3}$ \quad
Changqing Zou$^{1,5}$ \\
[0.2em]
$^{1}$ Zhejiang University \qquad
$^{2}$ Hong Kong Polytechnic University \\
$^{3}$ Nanyang Technological University
$^{4}$ University of the Chinese Academy of Sciences
$^{5}$ Zhejiang Lab\\
{\small $^{*}$Equal contribution.}
}

\newif\ifwithappendix
\withappendixfalse   
\ifwithappendix\else
  \makeatletter
  \AtBeginDocument{%
    \newlabel{app:imple_detail}{{A}{}{}{}{}}%
    \newlabel{app:complete_ablation}{{B}{}{}{}{}}%
    \newlabel{app:add_vis}{{C}{}{}{}{}}%
  }
  \makeatother
\fi
\begin{document}
\maketitle
\begin{abstract}
Learning how actions change visual states is essential for prediction and decision-making in embodied tasks.
Existing action-conditioned JEPA world models typically predict future representations at discrete observation intervals, leaving the temporal evolution of action effects implicit. 
We introduce ActField, a continuous-time JEPA world model that explicitly models latent state evolution through an action-conditioned vector field integrated over physical time. 
ActField separates action-independent evolution from action-induced changes, with the latter modulated by the latent context. 
A reference-conditioned representation and multi-step dynamics loss further support accurate long-horizon prediction.
Together, these components provide a structured and temporally continuous formulation for predicting action-conditioned visual dynamics.
Experiments across LIBERO, DROID, and BEHAVIOR demonstrate improved visual prediction fidelity and more accurate action-induced motion prediction over existing world models.
Further evaluations on simulated and real-world manipulation tasks demonstrate the utility of ActField in improving downstream policy planning through VLM-guided action selection.
\end{abstract}    

\begin{figure}
    \centering
    \includegraphics[width=\linewidth]{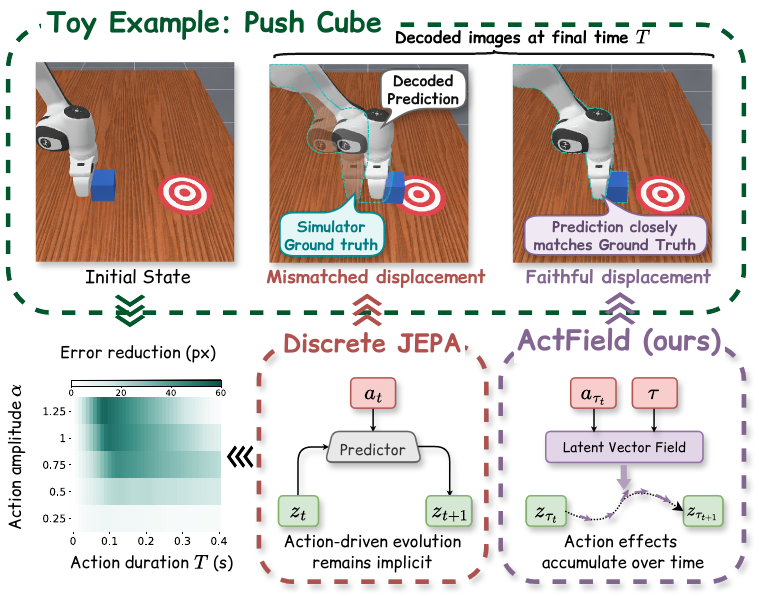}
    \caption{
        Comparison of world model predictions in a controlled cube-pushing task.
        Starting from the same initial state, we compare the decoded predictions of Discrete JEPA and ActField at the final time $T$.
        The heatmap reports how much ActField reduces cube-displacement error relative to Discrete JEPA across varying action amplitudes and execution durations, with both errors measured against the simulator outcome.
    }
    \label{fig:motivation}
\end{figure}
\section{Introduction}
\label{sec:introduction}

Robotic manipulation~\citep{intelligence2025pi_, qing2026pf, lin2026dygro} involves complex interactions between the embodied agent and its surrounding environments.
Predicting how these interactions change the environment~\citep{hafner2020mastering,finn2016unsupervised,finn2017deep} provides an effective means of assessing the downstream consequences of actions before execution.
World models~\citep{matsuo2022deep,ding2025understanding,hou2026world,ha2018recurrent} provide a foundation for this capability by learning how the environment evolves.
For robotic manipulation, an effective world model must accurately capture action-conditioned dynamics, as different action sequences from the same observation can produce distinct robot motions, contacts, and object displacements.

Recent advances in Joint-Embedding Predictive Architectures~(JEPA)~\citep{lecun2022path,assran2023self,cui2026generalization} offer a promising approach to visual world modeling by predicting target representations in latent space.
Representative action-conditioned JEPA models~\citep{zhou2024dino,maes2026leworldmodel,song2026actworld,liu2026toward} directly predict future visual representations from observation context and actions, demonstrating utility for downstream planning.
These models typically formulate latent dynamics as discrete transitions from the current context and the action executed over each observation interval to the next representation, recursively applying the predictor to imagine rollouts under action sequences.
We refer to this discrete-time latent prediction formulation as \emph{Discrete JEPA}.
Despite these promising results, this formulation leaves action-driven evolution within each transition interval implicit.
In particular, it does not explicitly represent how action-induced visual changes develop and accumulate during execution.
We illustrate this limitation in a controlled lateral cube-pushing task in ManiSkill~\citep{mu2021maniskill}, varying action magnitude and execution duration from the same physical state.
Figure~\ref{fig:motivation} presents both a representative visual case study and quantitative measurements of cube displacement.
Overlaying the final simulator configurations of the robot and block on the decoded prediction makes endpoint discrepancies visible.
Discrete JEPA exhibits a clear displacement mismatch in the visual example and fails to consistently track simulated block displacement across action magnitudes and execution durations.
Together, these results suggest that leaving action-driven evolution implicit limits the model's ability to predict action-dependent visual outcomes.
This motivates a central question:
\begin{tcolorbox}[notitle, rounded corners, colframe=darkgrey, colback=white, boxrule=2pt, boxsep=0pt, left=0.15cm, right=0.17cm, enhanced, shadow={2.5pt}{-2.5pt}{0pt}{opacity=5,mygrey},toprule=2pt, before skip=0.65em, after skip=0.75em]
\emph{
{\centering {\textbf{Can explicitly modeling action-driven latent evolution yield more faithful action-conditioned predictions?}}\\}
}
\end{tcolorbox}

In this paper, we propose \textbf{ActField}, an action-conditioned latent vector field model that predicts visual dynamics by accumulating local action-driven changes in latent space over time.
ActField maps pretrained visual features into a compact dynamic embedding conditioned on a shared reference observation, then evolves it using a velocity field that takes the reference embedding, current dynamic embedding, action, and physical time as inputs.
We structure the field as $f+G\psi(a)$, where $f$ captures evolution independent of the action input and $G$ transforms the action embedding $\psi(a)$ according to the current latent context.
Integrating the field over successive action intervals composes these local effects into future embeddings while allowing the action influence to vary with the evolving context and elapsed time.
We jointly supervise the visual feature reconstruction and integrated dynamics to align the learned latent evolution with observed visual changes.
As illustrated in Figure~\ref{fig:motivation}, the learned latent evolution more faithfully reflects the environment's underlying physical dynamics, yielding decoded predictions that better match simulator outcomes than Discrete JEPA.
For downstream decision-making, ActField predicts visual futures for candidate action sequences proposed by a behavior policy, and a vision-language model evaluates these predictions against the task instruction to select a sequence for execution.
Experiments across simulated and real-world manipulation tasks demonstrate consistent gains in both world-modeling accuracy and downstream decision-making performance.

Our main contributions are summarized as follows:
\begin{itemize}

    \item We identify implicit action-driven evolution as a key limitation of discrete-time action-conditioned JEPA models in faithfully capturing environment dynamics.
    
    \item We introduce \textbf{ActField}, a structured action-conditioned latent vector field that explicitly models and integrates local action effects over physical time.

    \item We show consistent gains in world-modeling accuracy and downstream decision-making through experiments across simulated and real-world manipulation tasks.
\end{itemize}
\section{Related Work}
\label{sec:related_work}
\subsection{Action-Conditioned Latent World Models}
Action-conditioned world models predict the consequences of control actions to support planning and policy learning~\citep{sekar2020planning,guo2026ctrl}, with latent approaches learning compact dynamics from visual observations~\citep{watter2015embed,zhang2019solar,hafner2019learning}.
Representative approaches, Dreamer~\citep{hafner2025mastering,hafner2023mastering, hafner2020mastering} learns policies from imagined trajectories generated by a latent world model, whereas TD-MPC~\citep{hansen2022temporal,hansen2024td,chitnis2024iql} uses learned latent dynamics to optimize action sequences at inference time through model-predictive control.
Joint-Embedding Predictive Architectures~(JEPA)~\citep{lecun2022path,mur2026v,chen2026vl} learn predictive representations by forecasting target embeddings rather than reconstructing pixels.
DINO-WM~\citep{zhou2024dino} learns an action-conditioned predictor over frozen DINOv2~\citep{oquab2023dinov2} features for latent-space planning.
V-JEPA 2-AC~\citep{assran2025v} scales JEPA pretraining on large-scale video and adapts it to action-conditioned robot modeling, while LeWorldModel~\citep{maes2026leworldmodel} stabilizes end-to-end pixel learning with a Sketched-Isotropic-Gaussian regularizer.
Despite these advances, action-conditioned JEPAs directly predict future representations from current representations and actions, leaving the intervening action-driven evolution implicit.
ActField instead models this evolution as an action-conditioned latent vector field integrated over physical time to obtain future representations.

\subsection{Continuous-Time Dynamics Modeling}
Continuous-time models describe state evolution through learned differential equations~\citep{zhang2025diffode, oh2025comprehensive, oh2024stable}.
Neural ODE~\citep{chen2018neural} models continuous state evolution using a neural-network-parameterized differential equation, while Latent ODE~\citep{rubanova2019latent} embeds this dynamics model into a latent-variable framework for modeling irregularly sampled time series.
More recently, continuous-time formulations have shown promise for embodied world modeling.
TAWM~\citep{nhu2025time} conditions latent dynamics on transition duration and trains across multiple temporal resolutions, enabling embodied prediction and control across varying observation frequencies.
ODEWorld~\citep{liu2026odeworld} further models visual dynamics as a physical-time latent velocity field and obtains future representations through numerical integration.
However, existing continuous-time world models have not explicitly focused on modeling how control actions induce the evolution of visual representations over physical time.
ActField addresses this gap by learning an action-conditioned latent vector field that directly models this state evolution.
\section{Preliminary}
\label{sec:preliminary}
\paragraph{Action-Conditioned JEPA World Models.}

Action-conditioned JEPA world models predict the consequences of control actions directly in a latent visual representation space, avoiding explicit reconstruction of future observations at the pixel level~\citep{zhou2024dino,assran2025v,zhang2026h}.
Given visual observations $o_t \in \mathcal O$ and control actions $a_t \in \mathcal A$, a visual encoder $\Phi_\phi$ maps each observation to a latent visual representation $h_t = \Phi_\phi(o_t)$.
Rather than predicting future pixels, an action-conditioned predictor $F_\theta$ models the transition between latent representations conditioned on the executed action, $\hat h_{t+1} = F_\theta(h_t, a_t)$.
The predictor directly parameterizes the transition from the current latent representation to that of the subsequent observation.
\begin{equation}
    \mathcal L_{\mathrm{JEPA}}
    =
    \mathbb E_{(o_t,a_t,o_{t+1})}
    \left[
        \mathcal D
        \left(
            F_\theta(h_t,a_t),
            \operatorname{sg}(h_{t+1})
        \right)
    \right],
    \label{eq:jepa_loss}
\end{equation}
where $\mathcal D(\cdot,\cdot)$ denotes a feature-space prediction loss and $\operatorname{sg}(\cdot)$ denotes stop-gradient on the target representation.
For multi-step prediction, the learned transition model is recursively applied along an action sequence $\mathbf a_{t:t+H-1}$ of horizon \(H\), where each predicted latent representation serves as the input to the subsequent transition.
This autoregressive rollout produces a sequence of future latent states without requiring a pixel-space decoder, providing the basis for latent-space prediction and planning.
\begin{figure*}[t] 
    \centering
      \includegraphics[width=0.95\textwidth]{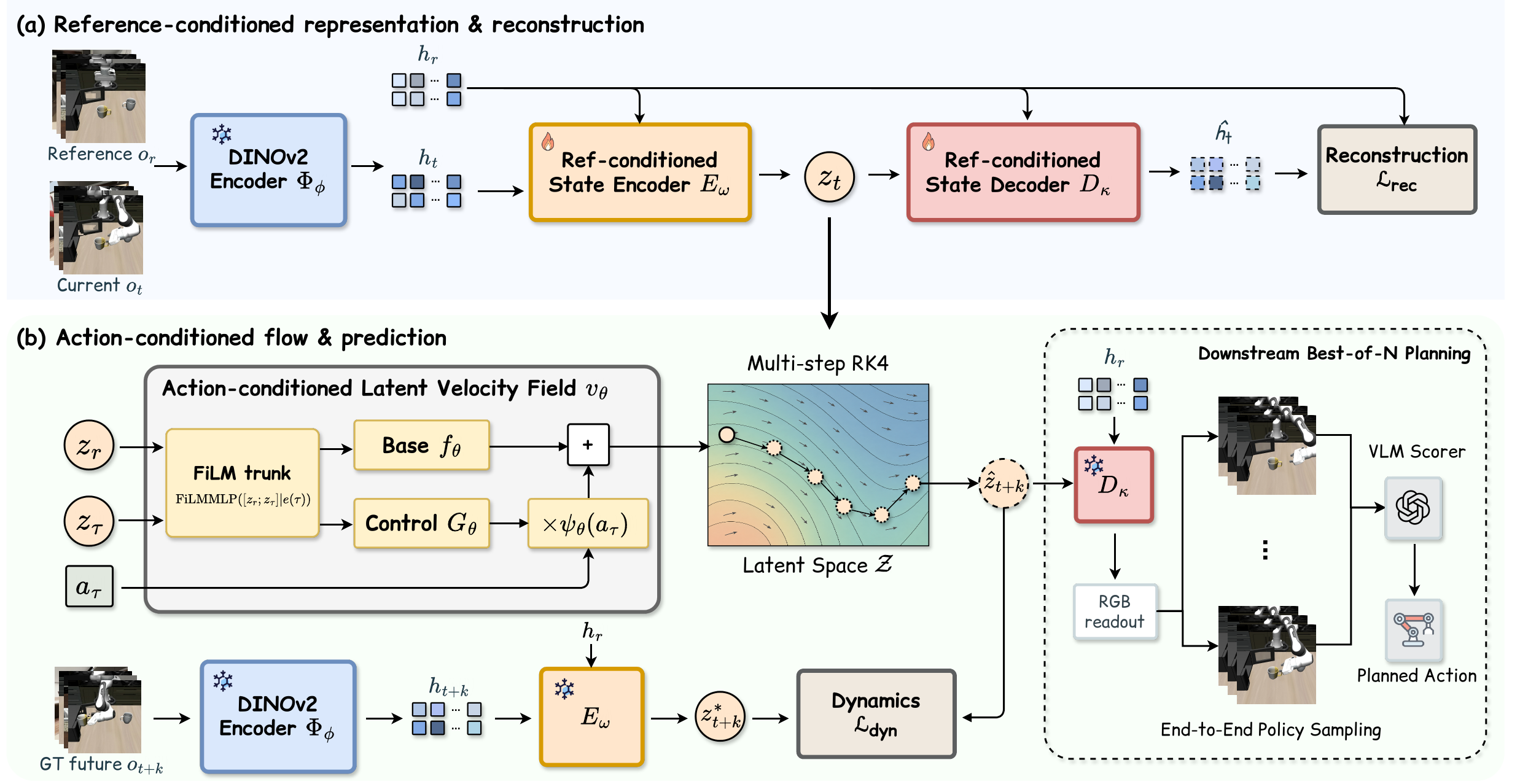} 
    \caption{An illustrative diagram of our proposed \textit{ActField}.}
    \vspace{-10pt}
    \label{fig:framework}
\end{figure*}
\section{Methodology}
\label{sec:method}

We introduce \textbf{ActField}, an action-conditioned JEPA world model that represents visual dynamics as a continuous-time latent vector field. 
As described in Sec.~\ref{sec:preliminary}, conventional action-conditioned JEPA models directly predict the next timestep visual representation from the current representation and control action. 
Our ActField instead introduces a compact reference-conditioned latent state and explicitly models how this state evolves under control actions over continuous physical time.
Specifically, given a reference observation, a current observation, and an input action sequence, ActField constructs a reference-conditioned latent state, explicitly models its local evolution with an action-conditioned vector field, and integrates the field over time to obtain future representations.
The same dynamics can be recursively integrated over successive control action steps to produce longer-horizon visual predictions.
An overview of ActField is shown in Figure~\ref{fig:framework}.

\subsection{Action-Conditioned Latent Flow Dynamics}
\label{sec:method:arch}

\paragraph{Reference-Conditioned State Encoding.}
Following Sec.~\ref{sec:preliminary}, let $o_t \in \mathcal O$ denote the visual observation at timestep $t$. 
We first employ the DINOv2~\citep{oquab2023dinov2} encoder \(\Phi_\phi:\mathcal O\rightarrow\mathbb{R}^{n\times d}\) to extract the patch-level visual representation \(h_t=\Phi_\phi(o_t)\).
The encoder remains frozen throughout ActField training, providing a fixed visual representation space for dynamics modeling.
Building on this fixed representation space, we further introduce a reference-conditioned dynamic encoder to extract compact states tailored for temporal evolution.
Specifically, given a reference observation $o_r$ from the same trajectory, we obtain
$h_r=\Phi_\phi(o_r)$ and use a dynamic encoder $E_\omega: \mathbb{R}^{n\times d} \rightarrow \mathcal Z$ to construct a compact reference-conditioned latent embedding:
\begin{equation}
    z_t = E_\omega(h_r,h_t).
    \label{eq:dynamic_state}
\end{equation}
We further define a fixed reference embedding \(z_r = E_\omega(h_r,h_r)\), which conditions the subsequent latent dynamics. 
The reference \(h_r\) provides a shared visual context for the trajectory, such that the current state and its future predictions are represented under the same reference condition.
Rather than directly evolving the high-dimensional visual representation $h_t$, our ActField performs dynamics modeling in the compact latent space $\mathcal Z$.

Meanwhile, to project the compact dynamic embeddings back into the shared visual representation space and facilitate downstream prediction evaluation, we further introduce a dynamic decoder $D_\kappa$ parameterized with \(\kappa\):
\begin{equation}
    \tilde h_t = D_\kappa(h_r,z_t).
    \label{eq:dynamic_decoder}
\end{equation}
Together, the reference-conditioned encoder and decoder define a compact dynamic state space while maintaining correspondence with the pretrained visual representation space.
Importantly, \(D_\kappa\) reconstructs visual representations rather than RGB pixels, keeping the world model entirely in the latent feature space.

\paragraph{Action-Conditioned Latent Velocity Field.}
To model how the latent embedding evolves under control actions, ActField treats \(z\) as a continuous-time variable. 
We use \(t\) to index discrete observation and control timesteps, while \(\tau\) denotes continuous physical time.
ActField models the latent evolution through an ordinary differential equation:
\begin{equation}
    \frac{\mathrm d z(\tau)}{\mathrm d\tau}
    =
    v_{\theta}
    \left(
        z_r,
        z(\tau),
        a(\tau),
        \tau
    \right),
    \label{eq:actfield_ode}
\end{equation}
where $v_{\theta}$ denotes our proposed action-conditioned latent vector field specifying the instantaneous rate of change of the latent embedding under the applied control signal $a(\tau)$.

To explicitly parameterize the contribution of action to the environment dynamics, we formulate the vector field as
\begin{equation}
    v_{\theta}(z_r,z,a,\tau)
    =
    f_{\theta_f}(z_r,z,\tau)
    +
    G_{\theta_G}(z_r,z,\tau)
    \psi_{\theta_a}(a),
    \label{eq:actfield_field}
\end{equation}
where $\theta=\{\theta_f,\theta_G,\theta_a\}$ denotes the parameters of the vector field.
\(f_{\theta_f}\) models the underlying latent evolution without explicit action influence, \(\psi_{\theta_a}\) maps the input action to control coefficients, and \(G_{\theta_G}\) specifies the latent directions and state-dependent influence associated with each control component. Their product \(G_{\theta_G}\psi_{\theta_a}(a)\) therefore represents the action-induced contribution to the local latent velocity.
This formulation explicitly parameterizes how actions affect the local latent dynamics, while future representations are obtained by integrating these dynamics over time rather than predicted directly.
We implement $f_{\theta_f}$ and $G_{\theta_G}$ as separate heads on a shared MLP trunk, where FiLM~\citep{perez2018film} modulates features of $[z_r; z]$ using a time embedding of $\tau$.
We also use a bias-free linear mapping for $\psi_{\theta_a}$.
For a fixed latent embedding and time, the action-dependent velocity therefore changes linearly with the encoded action.

\paragraph{Physical-Time Integration and Multi-Step Rollout.}
Given the action-conditioned vector field, ActField predicts future latent embeddings by integrating the continuous-time dynamics over corresponding physical time.
For an action sequence \(\mathbf a_{t:t+H-1}=(a_t,\ldots,a_{t+H-1})\), we assume zero-order hold and define the control signal as: 
\begin{equation}
    a(\tau)=a_{t+h},
    \qquad
    \tau\in[\tau_{t+h},\tau_{t+h+1}),
    \quad h=0,\ldots,H-1.
    \label{eq:zero_order_hold}
\end{equation}
Starting from the current latent embedding \(z_t\), ActField obtains the corresponding future embedding prediction at horizon \(H\) by integrating the learned vector field over the corresponding physical duration:
\begin{equation}
    \hat z_{t+H}
    =
    z_t+
    \int_{\tau_t}^{\tau_{t+H}}
    v_{\theta}
    \left(
        z_r,z(\tau),a(\tau),\tau
    \right)
    \mathrm d\tau.
    \label{eq:actfield_integral}
\end{equation}
We solve Eq.~\ref{eq:actfield_integral} using a differentiable Runge--Kutta integrator throughout training and inference.
During numerical integration, the vector field is repeatedly evaluated at intermediate latent embeddings generated along the predicted trajectory, without requiring explicit supervision on these intermediate states.
Thus, future representations are produced by temporally composing local action-conditioned dynamics, rather than directly predicting the overall transition from the current observation to a future one.
The predicted latent embedding is mapped back to the shared visual representation space through \(D_\kappa\):
\begin{equation}
    \hat h_{t+h}
    =
    D_\kappa(h_r,\hat z_{t+h}),
    \label{eq:rollout_decode}
\end{equation}
This yields a sequence of future visual representations conditioned on the prescribed action sequence.

\subsection{Representation and Dynamics Objectives}
\label{sec:method:objective}

In this section, we formulate the representation and dynamics objectives used to jointly train the ActField world model.
Let \(\mathcal D=\{(o_r,o_t,a_t,o_{t+1})\}\) denote the transition dataset.

\paragraph{Representation Objective.}
Since ActField models dynamics in the compact latent space $\mathcal Z$, we require the latent embeddings to remain grounded in the fixed pretrained visual representation space.
Given a reference observation $o_r$ and a current observation $o_t$, we obtain
\(h_r=\Phi_\phi(o_r)\) and \(h_t=\Phi_\phi(o_t)\) using the frozen DINOv2 visual encoder.
The dynamic encoder \(E_\omega\) constructs the reference-conditioned latent embedding \(z_t=E_\omega(h_r,h_t)\), which is decoded back to the visual feature space through \(D_\kappa\).
We therefore impose the following reconstruction objective
\begin{equation}
    \mathcal L_{\mathrm{rec}}(\omega,\kappa)
    =
    \mathbb E_{\mathcal D}
        \left\|
            D_\kappa(h_r,z_t)-h_t
        \right\|_2^2
    .
    \label{eq:reconstruction_loss}
\end{equation}
This objective encourages the compact latent embedding to retain the essential visual content of the current observation and organize it in a form that can be decoded to reconstruct the corresponding pretrained visual representation.
Since $\Phi_\phi$ remains frozen, $h_t$ provides a fixed feature-space target that anchors the evolving latent representation during dynamics learning.
The dynamic encoder $E_\omega$ and decoder $D_\kappa$ are optimized jointly with the action-conditioned velocity field, allowing the latent representation and the learned dynamics to co-adapt during training.

\paragraph{Dynamics Objective.}
We learn the action-conditioned velocity field $v_\theta$ from trajectory segments $(o_r,o_t,a_{t:t+h-1},o_{t+h})$ over multiple rollout horizons $h$.
For each horizon, we encode the future visual representation $h_{t+h}=\Phi_\phi(o_{t+h})$ under the same reference $h_r$ to construct the latent target: \(z_{t+h}^{\star}=\operatorname{sg}\left[E_\omega(h_r,h_{t+h})\right]\).
Using the same reference $h_r$ places the embedding $z_t$ and target $z_{t+h}^{\star}$ in a shared reference-conditioned latent space.

Starting from $z_t$, we obtain the predicted endpoint
$\hat z_{t+h}$ by integrating $v_\theta$ over the next $h$
action intervals according to Eq.~\ref{eq:actfield_integral}.
The integration follows the observed action sequence
$a_{t:t+h-1}$, with $a(\tau)=a_{t+j}$ for
$\tau\in[\tau_{t+j},\tau_{t+j+1})$ and $j=0,\ldots,h-1$.
Rather than directly supervising the instantaneous latent velocity,
we match the integrated latent displacement to the observed
displacement:
\begin{equation}
    \mathcal L_{\mathrm{dyn}}^{(h)}(\theta)
    \!=\!
    \mathbb{E}_{\mathcal D}
    \!\left\|
        \int_{\tau_t}^{\tau_{t+h}} \!
        v_{\theta}
        \left(
            z_r,\hat z(\tau),a(\tau),\tau
        \right)
        \mathrm d\tau
        -
        \left(
            z^\star_{t+h}\!-\!z_t
        \right)
    \right\|_2^2
    ,
    \label{eq:endpoint_loss}
\end{equation}
where $\hat z(\tau)$ is the predicted latent trajectory with $\hat z(\tau_t)=z_t$.
We optimize $v_\theta$ through differentiable
integration~\citep{chen2018neural}, averaging endpoint losses over horizons $\mathcal H=\{1,2,4,8\}$:
\begin{equation}
    \mathcal L_{\mathrm{dyn}}
    =
    \frac{1}{|\mathcal H|}
    \sum_{h\in\mathcal H}
    \mathcal L_{\mathrm{dyn}}^{(h)}.
    \label{eq:multistep_dynamics_loss}
\end{equation}
The complete ActField training objective is
\begin{equation}
    \mathcal L_{\mathrm{ActField}}
    =
    \mathcal L_{\mathrm{rec}}
    +
    \mathcal L_{\mathrm{dyn}}
    \label{eq:actfield_objective}
\end{equation}
During training, the visual encoder $\Phi_\phi$ remains frozen, while the dynamic encoder $E_\omega$, dynamic decoder $D_\kappa$, action mapping $\psi_{\theta_a}$, and velocity-field components
$f_{\theta_f}$ and $G_{\theta_G}$ are jointly optimized.
The stop-gradient in latent target computation is applied only to the future target branch, while the current embedding $z_t$ remains differentiable and receives gradients from both the representation and dynamics objectives.
Overall, ActField learns from observed visual transitions and their associated control actions without requiring instantaneous latent-velocity labels or intermediate trajectory annotations.
The resulting local dynamics can subsequently be composed over longer action sequences through physical-time integration.
\section{Experiments}
\label{sec:experiment}
To evaluate the effectiveness of ActField, we conduct experiments across multiple embodied benchmarks as well as real-world robotic tasks.
Our evaluation seeks to answer the following questions: 
(1)~Does ActField predict future visual observations more accurately than baseline world models over short and long horizons?~(Sections~\ref{sec:main_results} and~\ref{sec:visualization})
(2)~How accurately does ActField predict action-induced motion?~(Sections~\ref{sec:main_results}, \ref{sec:ablation}, and~\ref{sec:visualization})
(3)~Can ActField improve a fixed behavior policy's task success through VLM-guided action selection?~(Section~\ref{sec:main_results})
(4)~How do ActField's architectural components and training objectives contribute to its predictive performance?~(Section~\ref{sec:ablation})

\subsection{Experimental Setup}
We evaluate ActField on three benchmarks: LIBERO~\citep{liu2023libero}, DROID~\citep{khazatsky2024droid}, and BEHAVIOR~\citep{li2023behavior}.
We further assess its utility for action selection on two real-world manipulation tasks.
We compare ActField with representative JEPA-based world models, including DINO-WM~\citep{zhou2024dino} and LeWM~\citep{maes2026leworldmodel}.
Since these methods were originally evaluated under different experimental settings, we run all baselines using their official implementations under a common evaluation protocol.
Further implementation and evaluation details are provided in Appendix~\ref{app:imple_detail}.

\subsection{Main Results}
\label{sec:main_results}

\paragraph{Predictive fidelity.}
We evaluate visual prediction accuracy over short- and long-horizon open-loop rollouts, where each model recursively predicts future representations from an initial observation and action sequence without access to subsequent ground-truth observations.
We decode the predicted representations into RGB images using a pretrained DINOv2 decoder and evaluate their fidelity against the corresponding ground-truth frames using LPIPS~\citep{zhang2018unreasonable}, PSNR, and SSIM~\citep{wang2004image}.
Table~\ref{tab:visual_prediction} reports the results at $H=16$ and $H=64$. 
ActField consistently outperforms the baselines across all reported metrics and benchmarks, demonstrating its ability to maintain accurate and perceptually faithful visual predictions over extended horizons.

\begin{figure*}[t]
    \centering

    \includegraphics[width=0.42\textwidth]{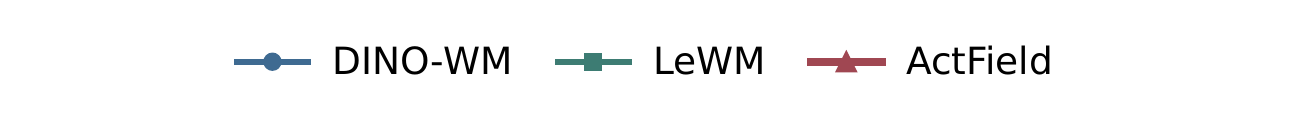}
    \vspace{2mm}

    \begin{subfigure}[t]{0.245\textwidth}
        \centering
        \includegraphics[width=\linewidth]{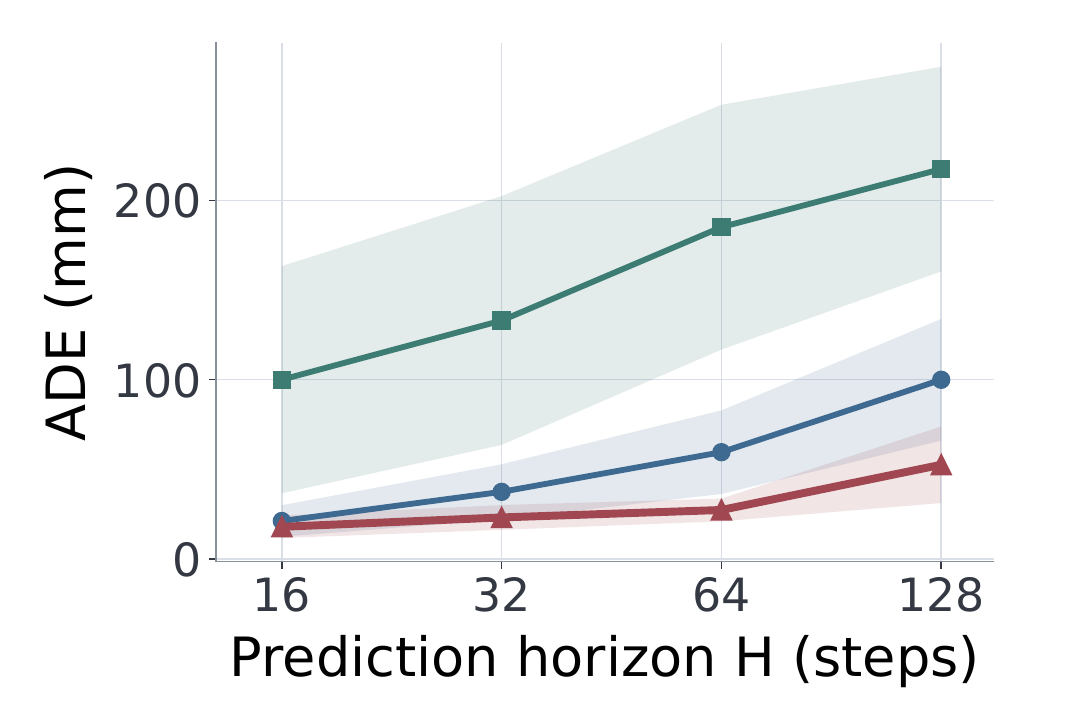}
        \caption{ADE over horizons.}
        \label{fig:eef_ade}
    \end{subfigure}
    \hfill
    \begin{subfigure}[t]{0.245\textwidth}
        \centering
        \includegraphics[width=\linewidth]{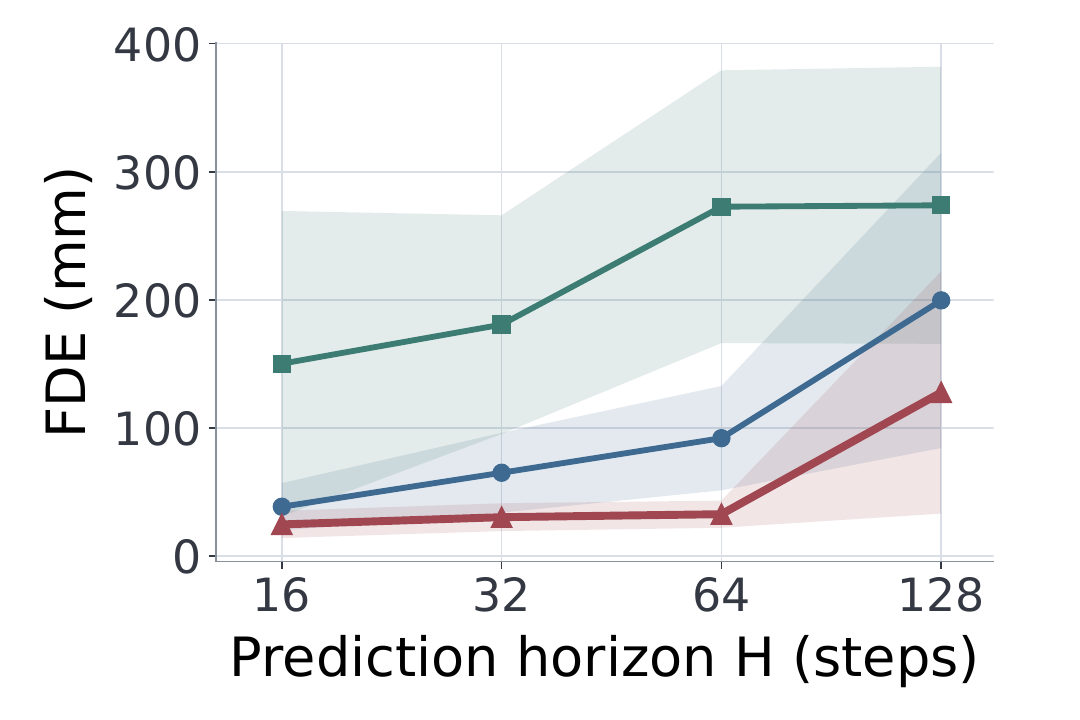}
        \caption{FDE over horizons.}
        \label{fig:eef_fde}
    \end{subfigure}
    \hfill
    \begin{subfigure}[t]{0.245\textwidth}
        \centering
        \includegraphics[width=\linewidth]{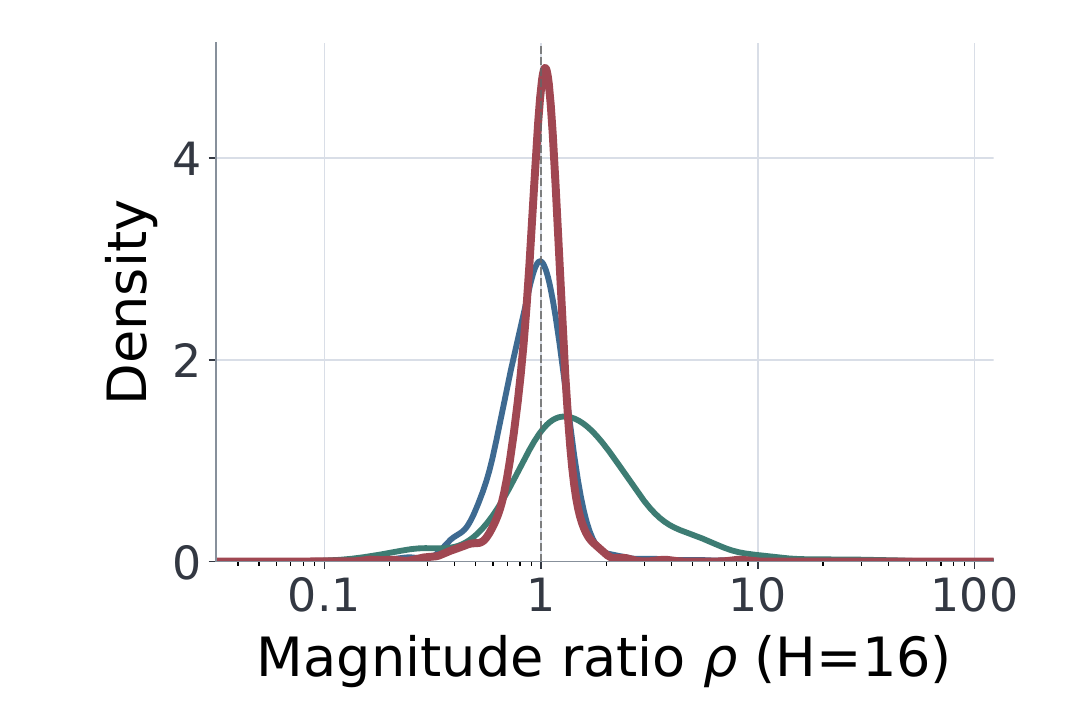}
        \caption{Magnitude ratio density.}
        \label{fig:eef_magnitude}
    \end{subfigure}
    \hfill
    \begin{subfigure}[t]{0.245\textwidth}
        \centering
        \includegraphics[width=\linewidth]{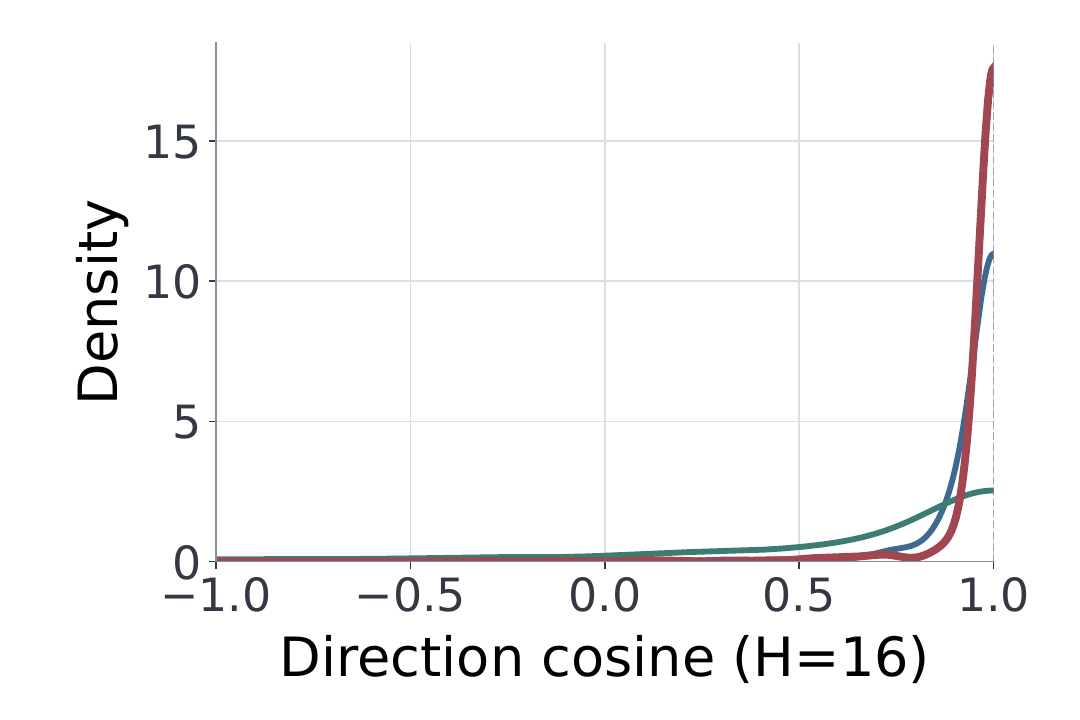}
        \caption{Directional similarity density.}
        \label{fig:eef_direction}
    \end{subfigure}

\caption{
    Physical motion from latent dynamics on LIBERO.
(a)~Average Displacement Error~(ADE) and (b)~Final Displacement
Error~(FDE) across horizons.
(c)~Displacement magnitude ratios and
(d)~Directional cosine similarities.
}
    \label{fig:eef_motion_prediction}
\end{figure*}

\begin{table*}[t]
    \centering
    \caption{Visual prediction quality of our ActField and other compared baselines at rollout horizons $H=16$ and $H=64$. All results use the 40,000-update checkpoint. Best and second-best results are marked in bold and underlined, respectively.}
    \label{tab:visual_prediction}
    \resizebox{\textwidth}{!}{
    \begin{tabular}{clccccccccc}
        \toprule
        \multirow{2}{*}{Horizon}
        & \multirow{2}{*}{Method}
        & \multicolumn{3}{c}{LIBERO~\citep{liu2023libero}}
        & \multicolumn{3}{c}{DROID~\citep{khazatsky2024droid}}
        & \multicolumn{3}{c}{BEHAVIOR~\citep{li2023behavior}} \\
        \cmidrule(lr){3-5}
        \cmidrule(lr){6-8}
        \cmidrule(lr){9-11}
        & & LPIPS$\downarrow$ & PSNR$\uparrow$ & SSIM$\uparrow$
        & LPIPS$\downarrow$ & PSNR$\uparrow$ & SSIM$\uparrow$
        & LPIPS$\downarrow$ & PSNR$\uparrow$ & SSIM$\uparrow$ \\
        \midrule

        \multirow{3}{*}{$H=16$} & DINO-WM~\citep{zhou2024dino}
        & $\underline{0.0530}$ & $\underline{26.399}$ & $\underline{0.8898}$
        & $\underline{0.0639}$ & $\underline{24.341}$ & $\underline{0.8607}$
        & $\underline{0.1852}$ & $\underline{20.627}$ & $\underline{0.6424}$ \\
         & LeWorldModel~\citep{maes2026leworldmodel}
        & $0.5580$ & $11.781$ & $0.4866$
        & $0.5708$ & $10.680$ & $0.3564$
        & $0.4833$ & $11.559$ & $0.3990$ \\
        \cmidrule(lr){2-11}
        \rowcolor{gray!15}
        \cellcolor{white}
        & \textbf{ActField}
        & $\mathbf{0.0395}$ & $\mathbf{27.769}$ & $\mathbf{0.9099}$
        & $\mathbf{0.0580}$ & $\mathbf{24.659}$ & $\mathbf{0.8706}$
        & $\mathbf{0.1701}$ & $\mathbf{20.632}$ & $\mathbf{0.6523}$ \\
        \midrule

        \multirow{3}{*}{$H=64$} & DINO-WM~\citep{zhou2024dino}
        & $\underline{0.1262}$ & $\underline{21.376}$ & $\underline{0.8150}$
        & $\underline{0.1197}$ & $\underline{21.027}$ & $\underline{0.8001}$
        & $\underline{0.2800}$ & $\underline{18.088}$ & $\underline{0.5855}$ \\
         & LeWorldModel~\citep{maes2026leworldmodel}
        & $0.5666$ & $11.159$ & $0.4650$
        & $0.5675$ & $10.594$ & $0.3489$
        & $0.4826$ & $11.733$ & $0.3987$ \\
        \cmidrule(lr){2-11}
        \rowcolor{gray!15}
        \cellcolor{white}
        & \textbf{ActField}
        & $\mathbf{0.0611}$ & $\mathbf{24.558}$ & $\mathbf{0.8758}$
        & $\mathbf{0.0786}$ & $\mathbf{22.422}$ & $\mathbf{0.8437}$
        & $\mathbf{0.2580}$ & $\mathbf{18.375}$ & $\mathbf{0.5939}$ \\
        \bottomrule
    \end{tabular}
    }
\end{table*}
\paragraph{Physical dynamics probing.}
Visual fidelity alone does not indicate whether predicted representations accurately capture underlying action-induced physical dynamics.
We train a separate MLP End-EFfector~(EEF) probe for each model using its ground-truth observation features, with matched architectures and training budgets, and evaluate the frozen probes on predicted features.
Specifically, let $\hat{\mathbf p}_{t+h}$ and $\mathbf p_{t+h}$ denote the predicted and ground-truth EEF positions at step $t+h$, respectively.
We measure trajectory accuracy using Average Displacement Error
$\mathrm{ADE}_H=\operatorname{mean}_h\|\hat{\mathbf p}_{t+h}-\mathbf p_{t+h}\|_2$
and Final Displacement Error
$\mathrm{FDE}_H=\|\hat{\mathbf p}_{t+H}-\mathbf p_{t+H}\|_2$.
As shown in Figures~\ref{fig:eef_ade} and~\ref{fig:eef_fde}, ActField consistently achieves lower errors across prediction horizons.
We further assess the magnitude and direction of predicted spatial EEF motion.
Defining $\widehat{\Delta\mathbf p}_H=\hat{\mathbf p}_{t+H}-\mathbf p_t$
and $\Delta\mathbf p_H=\mathbf p_{t+H}-\mathbf p_t$,
we compute the magnitude ratio
$\rho_H=\|\widehat{\Delta\mathbf p}_H\|_2/\|\Delta\mathbf p_H\|_2$
and directional similarity
$s_H^{\mathrm{dir}}=\cos(\widehat{\Delta\mathbf p}_H,\Delta\mathbf p_H)$.
Figures~\ref{fig:eef_magnitude} and~\ref{fig:eef_direction} show that ActField produces magnitude ratios more concentrated around 1 and directional similarities closer to 1, indicating more accurate predictions of both motion magnitude and direction.
Together, these results demonstrate that ActField accurately preserves physically meaningful action-conditioned dynamics beyond visual fidelity.

\begin{table*}[t]
    \centering
    \caption{
        Task success rates (\%) with VLM-guided best-of-$N$ action selection on
        (a) LIBERO-10 and (b) real-world manipulation tasks.
        All methods use the same behavior policy, candidate budget, and frozen VLM scorer.
        Results are based on 100 trials per task for LIBERO-10 and 20 trials per task for real-world manipulation. More details can be found in Appendix~\ref{app:imple_detail}.
    }
    \label{tab:task_success_all}

    \begin{subtable}[t]{0.63\textwidth}
        \vspace{0pt}
        \centering
        \caption{LIBERO-10.}
        \label{tab:task_success}
        \small
        \renewcommand{\arraystretch}{1.05}
        \setlength{\tabcolsep}{2pt}

        \begin{tabularx}{\linewidth}{@{}l*{11}{Y}@{}}
            \toprule
            \multirow{2}{*}{Rollout Model}
            & \multicolumn{10}{c}{Tasks}
            & \multirow{2}{*}{Avg.} \\
            \cmidrule(lr){2-11}
            & T1 & T2 & T3 & T4 & T5
            & T6 & T7 & T8 & T9 & T10
            & \\
            \midrule

            None (Base Policy)
            & 89 & 73 & 94 & 75 & 75
            & 95 & 68 & 86 & 34 & 28
            & 71.7 \\

            DINO-WM
            & 89 & 76 & 92 & 70 & 79
            & 94 & 62 & 88 & 27 & 18
            & 69.5  \\

            LeWorldModel
            & 82 & 64 & 84 & 62 & 66
            & 90 & 53 & 78 & 19 & 12
            & 61.0 \\

            \midrule
            \rowcolor{gray!15}
            ActField (Ours)
            & 94 & 83 & 93 & 84 & 84
            & 97 & 79 & 92 & 47 & 39
            & \textbf{79.2} \\

            \bottomrule
        \end{tabularx}
    \end{subtable}%
    \hfill
    \begin{subtable}[t]{0.35\textwidth}
        \vspace{0pt}
        \centering
        \caption{Real-world Tasks.}
        \label{tab:task_success_realworld}
        \small
        \renewcommand{\arraystretch}{1.15}
        \setlength{\tabcolsep}{2pt}
    
        \begin{tabularx}{\linewidth}{@{}lYYc@{}}
            \toprule
\multirow[c]{2}{*}{Rollout Model}
& Stack All
& Put Cubes
& \multirow[c]{2}{*}{Avg.} \\
& Plates
& into Bucket
            \\
            \midrule
    
            None (Base)
            & 60 & 55 & 57.5 \\
    
            DINO-WM
            & 60 & 50 & 55.0 \\
    
            LeWorldModel
            & 50 & 45 & 47.5 \\
    
            \midrule
            \rowcolor{gray!15}
            ActField (Ours)
            & 70 & 60 & \textbf{65.0} \\
    
            \bottomrule
        \end{tabularx}
    \end{subtable}
\end{table*}

\begin{figure}[t]
    \centering

    \includegraphics[width=0.33\columnwidth]{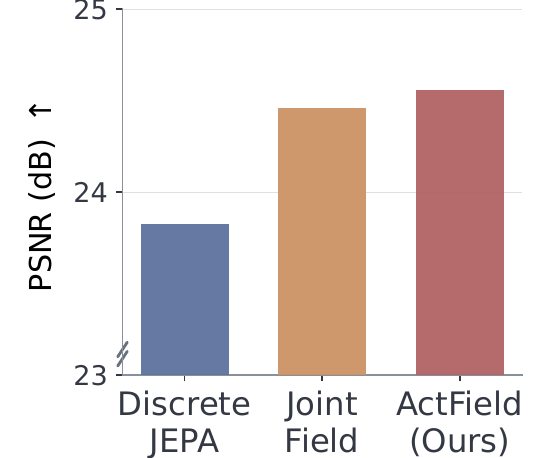}
    \hfill
    \includegraphics[width=0.32\columnwidth]{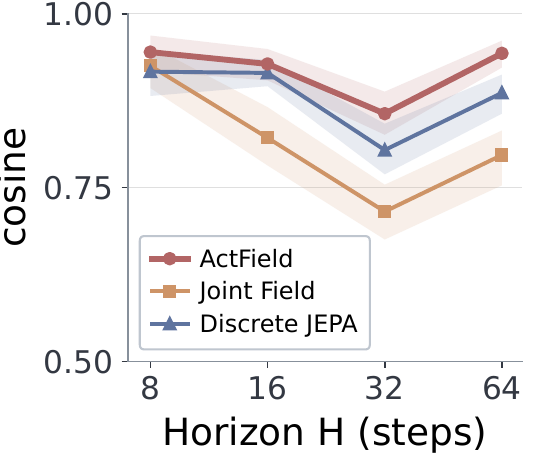}
    \hfill
    \includegraphics[width=0.33\columnwidth]{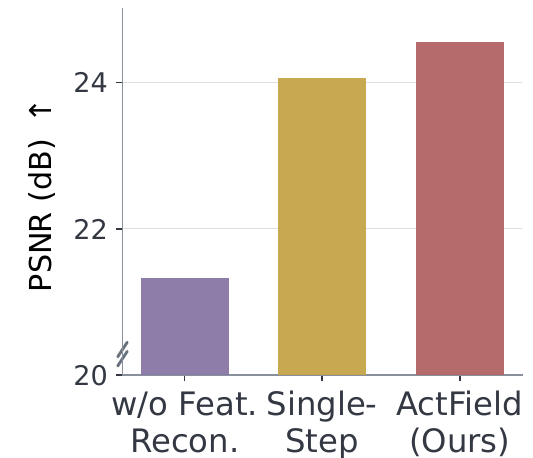}

    \caption{
        Ablation studies on LIBERO.
        Left:~Different dynamics parameterizations.
        Middle:~Action-effect cosine across prediction horizons.
        Right:~Contributions of feature reconstruction and multi-step dynamic loss at $H=64$.
        }
    \label{fig:ablation_main}
\end{figure}

\paragraph{Task-level policy enhancement.}
We further evaluate whether ActField's predictive capabilities translate into improved downstream task performance.
Following~\citet{alzayer2026masked}, we adopt a VLM-guided best-of-$N$ action-selection scheme.
At each decision step, a fixed stochastic behavior policy samples $N$ candidate action sequences, which are independently rolled out by each world model to predict future visual trajectories.
A shared Astra VLM scorer evaluates the predicted trajectories against the task instruction and selects the highest-scoring candidate for execution.
All methods use the same behavior policy, candidate budget, and VLM scorer, ensuring a controlled comparison of their decision utility.
Further details of the inference procedure and its configurations are provided in Appendix~\ref{app:imple_detail}.
Tables~\ref{tab:task_success} and~\ref{tab:task_success_realworld} report per-task and average success rates on LIBERO-10 and real-world manipulation tasks, respectively.
On LIBERO-10, ActField-guided action selection consistently improves upon the base policy and outperforms alternative world models.
The improvements also extend to real-world manipulation, demonstrating the practical utility of ActField for VLM-guided action selection beyond simulation.

\subsection{Ablation Studies}
\label{sec:ablation}

\begin{table}[t]
    \centering
    \caption{
        Effect of reference conditioning on visual prediction quality.
        LPIPS ($\downarrow$) at $H=64$ across three benchmarks.
    }
    \label{tab:reference_ablation}
    \small
    \setlength{\tabcolsep}{3pt}
    \renewcommand{\arraystretch}{1.05}
    \begin{tabular}{@{}lcccc@{}}
        \toprule
        Method
        & LIBERO
        & DROID
        & BEHAVIOR
        & Avg. \\
        \midrule
        w/o Reference
        & 0.1083 & 0.1577 & 0.3697 & 0.2119 \\
        \rowcolor{gray!15}
        ActField (Ours)
        & \textbf{0.0611} & \textbf{0.0786} & \textbf{0.2580} & \textbf{0.1326} \\
        \bottomrule
    \end{tabular}
\end{table}

\paragraph{Training objectives.}
We investigate the individual and combined contributions of feature reconstruction and the multi-step dynamics loss to predictive performance.
As shown in the right panel of Figure~\ref{fig:ablation_main}, replacing the multi-step dynamics loss with single-step endpoint supervision~(Single Step) leads to a reduction in PSNR, indicating that one-step prediction alone is insufficient to maintain accuracy over recursive rollouts.
This highlights the importance of supervising the integrated dynamics across multiple prediction horizons to mitigate error accumulation during long-horizon prediction.
Removing the feature reconstruction loss also degrades performance, highlighting its role in anchoring predicted dynamic embeddings to the frozen visual representation space.
By constraining the integrated predictions in the shared visual feature space, this objective helps preserve visual consistency throughout latent evolution.
The full model achieves the best performance, demonstrating that multi-step dynamics learning and feature reconstruction provide complementary benefits for accurate and visually consistent long-horizon prediction.

\paragraph{Dynamics parameterization.}
We investigate the benefits of the structured $f+G\psi$ parameterization from both visual prediction and action-effect estimation perspectives.
We compare ActField with two alternative dynamics formulations.
Discrete JEPA directly predicts the next latent representation from the current representation and action, while Joint Field parameterizes the complete latent velocity using a single action-conditioned network without the decomposition in Eq.~\ref{eq:actfield_field}.
As shown in the left panel of Figure~\ref{fig:ablation_main}, Joint Field achieves higher PSNR than Discrete JEPA, highlighting the advantage of continuous-time latent dynamics over discrete transitions.
ActField further improves PSNR, demonstrating the benefit of explicitly separating autonomous and action-induced components in the vector field.
Beyond visual prediction, we examine whether this decomposition improves the representation of action-induced physical effects.
We estimate object displacements from predicted latent changes and measure their cosine similarity with the corresponding simulator displacements.
As shown in the middle panel of Figure~\ref{fig:ablation_main}, ActField achieves higher cosine similarity than Joint Field, with a more pronounced advantage at longer prediction horizons.
This indicates that the structured parameterization better preserves the physical information of action-induced effects during long-horizon prediction, supporting the effectiveness of explicitly modeling action contributions through $G$.

\paragraph{Reference conditioning.}
We further examine the role of reference conditioning by removing the reference observation from both the dynamics encoder and decoder.
As reported in Table~\ref{tab:reference_ablation}, removing reference conditioning consistently increases LPIPS across LIBERO, DROID, and BEHAVIOR, with the average LPIPS increasing from 0.1326 to 0.2119.
This degradation suggests that conditioning on a reference observation provides a stable representation of scene-specific visual context, allowing the dynamic embeddings to focus on task-specific action-induced changes rather than repeatedly encoding static scene information.
The reference therefore serves as a shared visual anchor for both latent dynamics modeling and feature reconstruction, improving the consistency of predicted representations over long-horizon rollouts.
These results highlight the importance of reference conditioning in learning accurate and visually consistent latent dynamics across diverse environments.
Further ablation results are provided in Appendix~\ref{app:complete_ablation}.

\begin{figure*}[t]
    \centering
    \includegraphics[width=\textwidth]{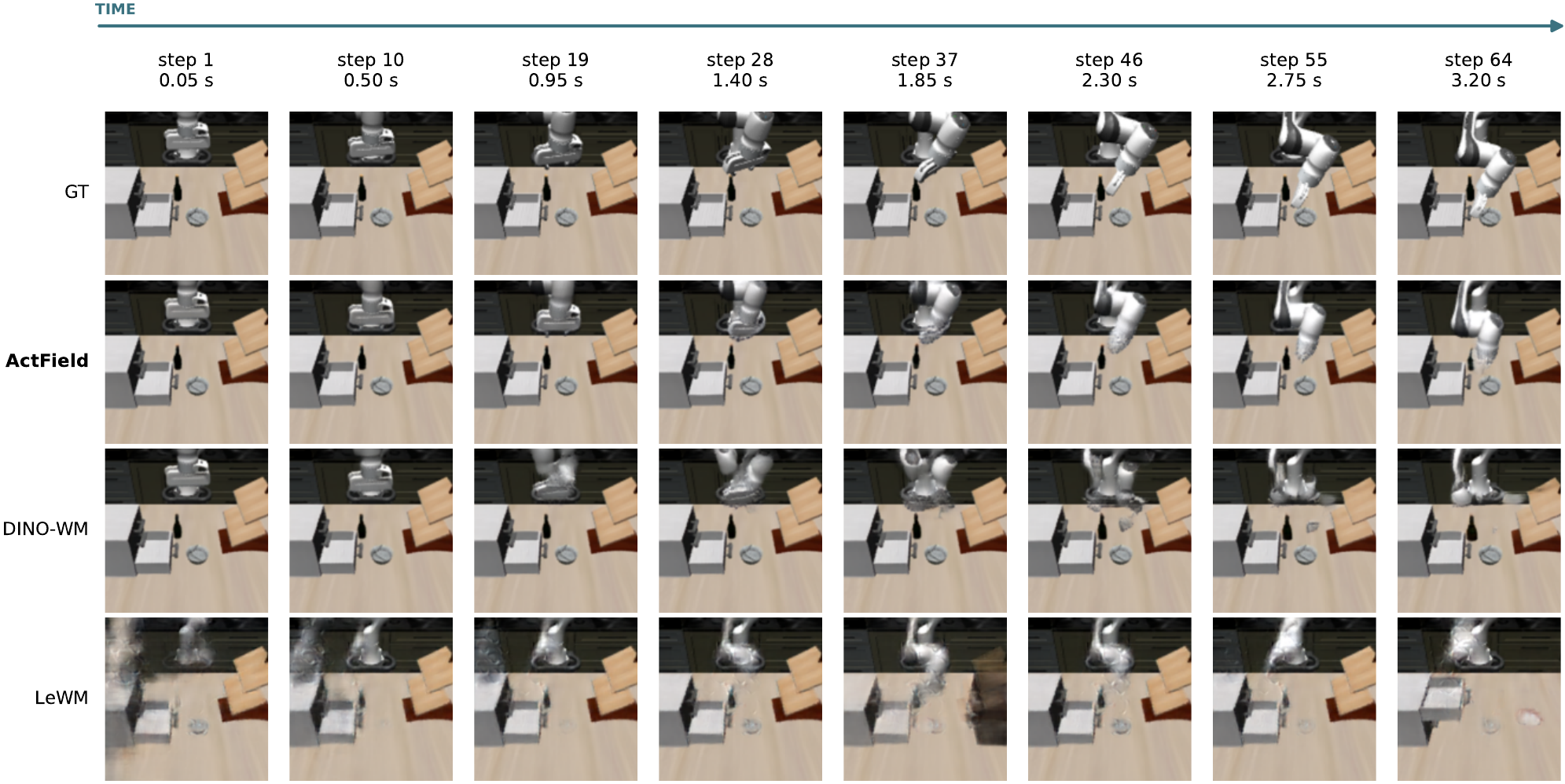}
    \caption{
    64-step visual rollouts on LIBERO.
    Ground truth (GT) denotes simulator observations obtained by executing the action sequence.
    The remaining rows show decoded model predictions conditioned on the same initial observation and actions.
    }
    \label{fig:visualization_reconstruction}
\end{figure*}

\begin{figure}[t]
    \centering
    \includegraphics[width=0.99\linewidth]{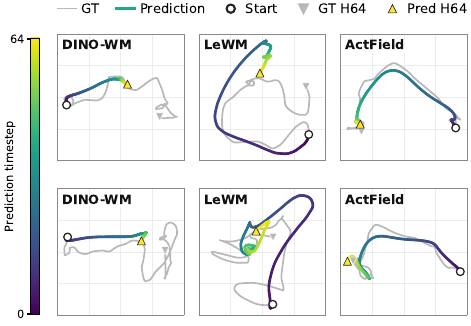}
    \caption{
    PCA projections of 64-step latent rollout trajectories.
The top and bottom rows show two representative manipulation tasks from LIBERO-Goal: \emph{push the plate to the front of the stove} and \emph{put the bowl on the plate}, respectively.
    }
    \label{fig:visualization_features}
\end{figure}

\begin{figure}[t]
    \centering

    \includegraphics[width=0.7\linewidth]
        {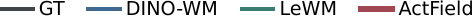}
    \vspace{0ex}

    \includegraphics[width=0.32\linewidth]
        {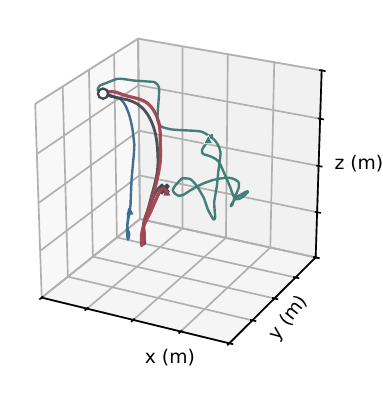}
    \hfill
    \includegraphics[width=0.32\linewidth]
        {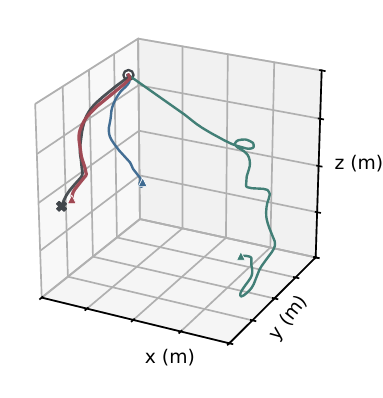}
    \hfill
    \includegraphics[width=0.32\linewidth]
        {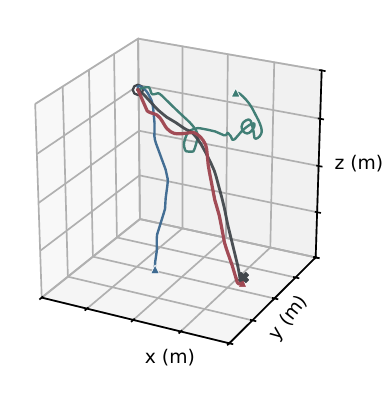}

    \caption{
        Probed end-effector trajectories over 64 rollout steps.
        We compare predicted 3D trajectories with ground truth on three LIBERO manipulation tasks.
        The task instructions are, from left to right,
        \emph{put the bowl on the plate},
        \emph{put the alphabet soup and tomato sauce in the basket}, and
        \emph{put the moka pot on the stove}.
    }
    \label{fig:visualization_eef}
\end{figure}
\subsection{Visualization}
\label{sec:visualization}
To further examine the predicted dynamics, we visualize decoded observations, latent feature trajectories, and end-effector motion over 64-step open-loop rollouts.
Figure~\ref{fig:visualization_reconstruction} compares decoded predictions with ground-truth observations at selected timesteps.
ActField better preserves visual details and follows the evolving robot poses throughout the rollout, while DINO-WM and LeWorldModel exhibit more noticeable deviations.
Figure~\ref{fig:visualization_features} compares PCA projections of latent rollout trajectories with ground-truth feature evolution on two LIBERO-Goal tasks.
ActField more closely follows the ground-truth trajectories in both tasks, whereas the baselines exhibit larger deviations in trajectory shapes and endpoints.
Finally, Figure~\ref{fig:visualization_eef} shows the 3D end-effector trajectories decoded from predicted latent features using an MLP probe.
Across three manipulation tasks, ActField better preserves the spatial paths and final positions of the ground-truth trajectories.
Together, these visualizations demonstrate that ActField better captures visual, latent, and physical dynamics over extended rollouts.
More visualization results are provided in Appendix~\ref{app:add_vis}.

\section{Conclusion}
\label{sec:conclusion}
In this work, we propose \textbf{ActField}, a continuous-time JEPA world model that explicitly models action effects through a latent vector field integrated over physical time. By decomposing latent dynamics into autonomous and action-induced components, ActField captures how control actions shape latent state evolution. A reference-conditioned representation and multi-step dynamics loss further support accurate long-horizon prediction. Experiments across LIBERO, DROID, and BEHAVIOR demonstrate improved visual prediction fidelity over existing world models, while physical dynamics probing and ablation studies validate the proposed design. Furthermore, evaluations on VLM-guided action selection in simulation and real-world manipulation demonstrate the practical utility of ActField for predictive decision-making in embodied systems.
\section*{LLM Usage Statement}

We used GPT-6 to assist with software programming and manuscript language polishing and editing. We also used Astra as a VLM scorer to evaluate candidate trajectories against task instructions in our action-selection experiments. These models were not used for research idea generation or experimental data synthesis. All AI-assisted text was reviewed and revised by the authors, and all AI-assisted code was inspected and tested for correctness. The authors take full responsibility for the final content of this work.
{
    \small
    \bibliographystyle{ieeenat_fullname}
    \bibliography{main}
}

\end{document}